\documentclass[letterpaper]{article} % DO NOT CHANGE THIS
\usepackage{aaai2026}  % DO NOT CHANGE THIS
\usepackage{times}  % DO NOT CHANGE THIS
\usepackage{helvet}  % DO NOT CHANGE THIS
\usepackage{courier}  % DO NOT CHANGE THIS
\usepackage[hyphens]{url}  % DO NOT CHANGE THIS
\usepackage{graphicx} % DO NOT CHANGE THIS
\usepackage{natbib}  % DO NOT CHANGE THIS AND DO NOT ADD ANY OPTIONS TO IT
\usepackage{caption} % DO NOT CHANGE THIS AND DO NOT ADD ANY OPTIONS TO IT
\usepackage{booktabs}
\usepackage{multirow}
\usepackage{amsmath}
\usepackage{xspace}
\usepackage[table]{xcolor}
\usepackage{amssymb}
\usepackage{appendix}

\usepackage{algorithm}
\usepackage{algorithmic}

\usepackage{xcolor}
\usepackage{newfloat}
\usepackage{listings}
\DeclareCaptionStyle{ruled}{labelfont=normalfont,labelsep=colon,strut=off} % DO NOT CHANGE THIS
\floatstyle{ruled}
\newfloat{listing}{tb}{lst}{}
\floatname{listing}{Listing}
\title{Point Cloud Segmentation of Integrated Circuits Package Substrates Surface Defects Using Causal Inference: Dataset Construction and Methodology}
\author{
    Bingyang Guo,
    Qiang Zuo,
    Ruiyun Yu\thanks{Corresponding Author.}
}
\affiliations{
    Software College, Northeastern University, Shenyang, China \\
    2110500@stu.neu.edu.cn, 2271499@stu.neu.edu.cn, yury@mail.neu.edu.cn
}

\usepackage{bibentry}
\begin{document}

\maketitle

\begin{abstract}
The effective segmentation of 3D data is crucial for a wide range of industrial applications, especially for detecting subtle defects in the field of integrated circuits (IC). Ceramic package substrates (CPS), as an important electronic material, are essential in IC packaging owing to their superior physical and chemical properties. However, the complex structure and minor defects of CPS, along with the absence of a publically available dataset, significantly hinder the development of CPS surface defect detection. In this study, we construct a high-quality point cloud dataset for 3D segmentation of surface defects in CPS, i.e., \textbf{CPS3D-Seg}, which has the best point resolution and precision compared to existing 3D industrial datasets. CPS3D-Seg consists of 1300 point cloud samples under 20 product categories, and each sample provides accurate point-level annotations. Meanwhile, we conduct a comprehensive benchmark based on SOTA point cloud segmentation algorithms to validate the effectiveness of CPS3D-Seg. Additionally, we propose a novel 3D segmentation method based on causal inference (CINet), which quantifies potential confounders in point clouds through Structural Refine (SR) and Quality Assessment (QA) Modules. Extensive experiments demonstrate that CINet significantly outperforms existing algorithms in both mIoU and accuracy. 
\end{abstract}

\begin{links}
    \link{Code \& Datasets}{https://github.com/Bingyang0410/CPS3D-Seg}
\end{links}

\section{Introduction}
\label{sec:intro}
3D point cloud segmentation is vital in industrial condition monitoring, as it enables a comprehensive understanding of the geometric structure and intricate characteristics of each point. This capability is increasingly critical in the high-end manufacturing sector of integrated circuits, where precise defect detection and quality assurance are essential for maintaining reliability and performance. Ceramic package substrates, known for their exceptional material properties such as high hardness, elevated elastic modulus, extreme wear resistance, excellent electrical insulation, and superior corrosion resistance, play a crucial role in the integrated circuits industry by ensuring the reliability and performance of electronic devices \cite{Alias2010DefectOO}. However, manufacturing defect-free ceramic package substrates remains challenging, with common defects such as cracks, holes, bubbles, delamination, scratches, and stains \cite{9351787} potentially compromising substrate quality and increasing the likelihood of electronic equipment failure.

\begin{figure}[tp]
    \centering
    \includegraphics[width=\linewidth]{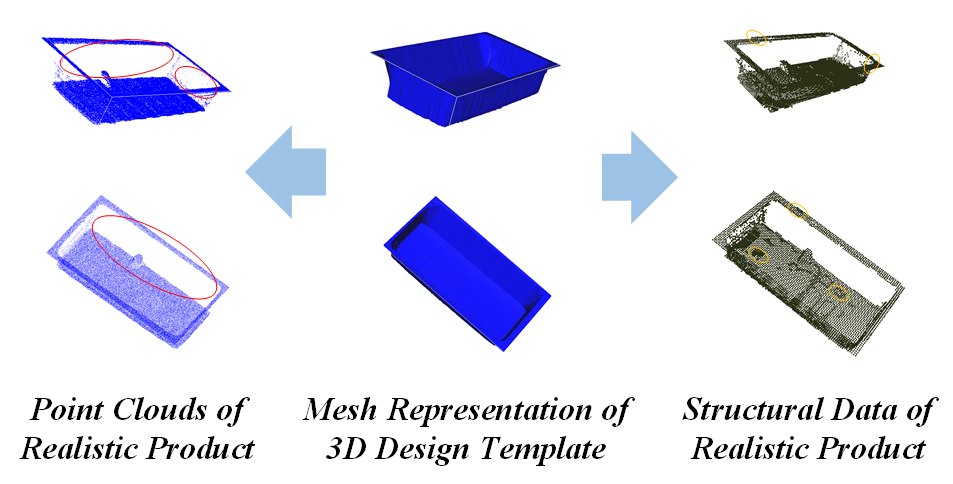}
    \caption{The demonstration of discrepancies in the presentation effect. The red oval box represents the missing stereo data during the scanning process, and the yellow oval box means the aggregation errors in the structuring process.}
    \label{fig:1}
\end{figure}

Existing research \cite{huang2022applying, huang2021imbalance, dahai2023lightweight} has made initial advances in detecting surface defects in substrates, primarily relying on 2D image-based methods. While these 2D techniques have shown some effectiveness, they inherently need more depth information, making it difficult to accurately represent circuits' vertical connectivity and the substrate surfaces' precise topography. This limitation impedes the reliable detection and analysis of defects lying in the depth dimension of the substrate. Recent advancements in 3D acquisition technologies, such as stereo cameras \cite{988771}, structured light \cite{SALVI20102666}, LiDAR \cite{10286126}, and laser scanners \cite{son2002automated}, have provided a promising solution to this problem by enabling the capture of complex 3D data. These technologies offer a comprehensive view of the ceramic package substrates from multiple perspectives, allowing for a more precise and detailed surface quality assessment. The integration of 3D data provides richer information, facilitating better identification and characterization compared to 2D data.

Although several 3D point cloud datasets \cite{DBLP:conf/mm/ChenXLWLWZ23,DBLP:conf/wacv/BergmannS23,DBLP:journals/corr/abs-2311-14897,DBLP:journals/corr/abs-2312-04521,DBLP:conf/visapp/BergmannJSS22,DBLP:conf/accv/BonfiglioliTSFG22,DBLP:conf/nips/LiuXCLWLWZ23} are currently available for defect detection, these datasets frequently exhibit substantial differences when compared to ceramic package substrates. Typical 3D datasets may include objects with relatively simple geometries and textures. In contrast, ceramic package substrates feature highly intricate surface circuitry and complex three-dimensional structures \cite{LIANG2023103856}. The point cloud density and accuracy required in integrated circuits are much higher, making data acquisition more challenging. More publicly available point cloud datasets tailored for integrated circuits are needed to address this issue in this field.

% \footnote{https://www.keyence.com/products/measure/laser-2d/lj-x8000/}

% \footnote{https://www.cloudcompare.org/}

To address the above problem, we present a high-resolution point cloud segmentation dataset of ceramic package substrates, named CPS3D-Seg. Unlike previous 3D datasets collected by RGB-D cameras or double CCD structured light, the point cloud data of CPS3D-Seg are all obtained by high-precision line laser scanners Keyence LJ-X8020. Specifically, we have built a data acquisition platform that integrates four LJ-X8020s, ensuring accuracy and data acquisition efficiency. Thanks to the improved performance of the collection equipment, the point cloud density and accuracy of the CPS3D-Seg dataset have significantly improved. CPS3D-Seg offers 16K point clouds per object with a point precision of 2.5 ${\mu}m$, which exceeds the current highest precision dataset Real3D-AD \cite{DBLP:conf/nips/LiuXCLWLWZ23} by an order of magnitude. What's more, we utilize CloudCompare to accurately annotate each defect sample at the point level, visually presenting concrete defect information.

Inspired by previous excellent datasets, we also construct a comprehensive and large-scale benchmark to evaluate the effectiveness of CPS3D-Seg. As part of the benchmark construction, we perform qualitative and quantitative analysis of the data and algorithms. Our research results indicate that the application of specific preprocessing or structuring procedures may result in discrepancies in the presentation effect due to the increased density and accuracy of point clouds. As shown in Fig. \ref{fig:1}, this can be summarized into two folds: i) The red oval box represents that the line laser may bring out information missing during the scanning process due to its angle of collection. The direct application of the point cloud completion approach will yield imprecise data. ii) The yellow oval box means that aggregation errors in the structuring process may combine certain defective and non-defective areas. By comparing these with the mesh representation of the 3D design template, the above issues become more apparent.

In essence, all the aforementioned concerns can be encapsulated as the intricacy of the point cloud. Inspired by this discovery, we propose an innovative approach utilizing causal intervention into the formulation of point cloud segmentation, which helps in discerning latent relationships of high-resolution 3D data. Technically, we conduct a structural causal model to explain the potential effect process. In response to the difficulty of measuring or observing the hidden confounder in this problem, we propose a causal intervention approach that relies on back-door adjustment. Our approach uses the CPS3D-Seg dataset for practical 'virtual' interventions, rather than overly expensive 'physical' interventions such as changing the collection angle or repeating the collection multiple times.

The primary contributions of this paper are as follows:

\begin{enumerate}
    \item We construct a 3D segmentation dataset suitable for ceramic package substrates surface defect detection, named CPS3D-Seg, which fills the gap in the field of integrated circuits and surpasses existing datasets in point cloud resolution and accuracy.
    \item We propose a benchmark for 3D segmentation on CPS3D-Seg, encompassing latest SOTA algorithms. This benchmark evaluates the effectiveness of CPS3D-Seg.
    \item We introduce a novel 3D segmentation method based on structural causal model, which achieves causal inference by solving probabilities through learnable modules. Comprehensive experimental results illustrate the superior efficacy of our approach.
\end{enumerate}

\section{Related Work}
\label{sec:related}

\textbf{3D Industrial Segmentation Datasets.} In recent years, the importance of point cloud has significantly increased in numerous fields, such as manufacturing \cite{Wu20143DSA}, autonomous driving \cite{Geiger2012CVPR,hackel2017isprs}, scene understanding \cite{Mo_2019_CVPR, 8099744,Silberman:ECCV12}, and others. Due to the privacy of industrial data, there are currently only a small number of industrial segmentation datasets available. MVTec 3D-AD \cite{DBLP:conf/visapp/BergmannJSS22} is the first public dataset in this field, and it contains over 4000 samples with single-view information obtained from Zivid One Plus. This dataset encompasses several industrial components, each category including typical items in actual manufacturing while offering a wide range of abnormalities. By using a handheld scanning device to collect $360^{\circ}$ data from the object, Real3D-AD \cite{DBLP:conf/nips/LiuXCLWLWZ23} enhances the integrity and accuracy of the point cloud. Moreover, Real3D-AD conducts a comprehensive benchmark which provides detailed experimental results and comparative references. In order to reduce the difficulty of data collection, Eyecan-ai published a synthetic 3D dataset with multiple multimodalities \cite{DBLP:conf/accv/BonfiglioliTSFG22}, such as such as RGB images, depth maps, and camera pose matrices. Anomaly-ShapeNet \cite{Li_2024_CVPR} offers an extensive synthetic 3D point dataset designed explicitly for detection tasks, significantly enhancing the sample size of the dataset. This dataset comprises a range of difficulty samples, offering researchers a foundation for assessing and contrasting different algorithms. However, significant constraints persist in the accuracy of point clouds and the authenticity of data sources within existing datasets, and there is no accessible 3D dataset to advance the study of integrated circuits.

\textbf{3D Segmentation Methods.} With the rapid development of 3D sensor technology, point cloud segmentation has played a crucial role in fields such as autonomous driving, robot navigation, and surface detection. Deep learning has achieved impressive performance improvements over traditional methods for 3D segmentation tasks. According to the different architectures of the backbone, these methods can be roughly divided into CNN-based methods, graph-based methods, transformer-based methods, and mamba-based methods. CNN-based methods \cite{8099499, 10.5555/3295222.3295263,Horwitz_2023_CVPR,wang2023multimodal,9879738,DBLP:conf/nips/LiuXCLWLWZ23,qian2022pointnext,10204778,peng2024oacnns} voxelize point cloud data to adapt to existing 2D and 3D convolutional structures, but it also brings about issues of computational complexity and memory consumption. Graph-based methods \cite{dgcnn,Wang2019_GACNet, 9156514,lei2020spherical,10.1109/TPAMI.2023.3238516,robert2024scalable} can directly capture local and global structural information in point cloud data, in order to more effectively capture the dependency relationships between points and achieve efficient segmentation. By introducing self-attention mechanism, transformer-based methods \cite{DBLP:conf/iccv/ZhaoJJTK21,DBLP:conf/nips/0002LJLZ22,wu2024ptv3,10543016,Wang2023OctFormer,kolodiazhnyi2024oneformer3d} can capture long-range dependencies of point cloud data on a global scale, enabling it to maintain efficiency and accuracy in processing large-scale point cloud data. Mamba-based methods \cite{zhang2024point,han2024mamba3d,liu2024point,liang2024pointmamba,Wang2024PoinTrambaAH} using state space model (SSM) have achieved both linear complexity and the long-range context learning abilities in point cloud data. However, the above methods do not have specific designs for complex ceramic package substrate data.

\textbf{Causal Inference.} 
The goal of causal inference \cite{ci} is to quantify the causal relationship between variables, not just the correlation of statistical data. There is an increasing prevalence of computer vision tasks \cite{DBLP:conf/ijcai/GrariLD22,DBLP:conf/iccv/YueS0Z21,DBLP:conf/kdd/RenCMWY22,DBLP:conf/cvpr/NiuTZL0W21,DBLP:conf/cvpr/QiNHZ20,DBLP:conf/iclr/ZhangGL000S022,DBLP:conf/nips/ZhangZT0S20} that leverage causal inference, enhancing traditional CNN's performance across multiple dimensions. To the best of our knowledge, limited research has employed causal inference in the field of 3D segmentation. D-S \cite{ZHANG2023104915} develops a comprehensive causal inference methodology to achieve multi-label segmentation of diverse objects from 3D point clouds of tunnels. CausalIPC \cite{Huang_2024_CVPR} has developed a quantifiable method to eliminate the influence of potential factors, substantially enhancing the adversarial robustness of the prominent point cloud classification method. However, the structural causal model designed by the above methods does not apply to 3D segmentation of ceramic package substrates.

\section{CPS3D-Seg Dataset}
\label{sec:data}
\textbf{Data source.} Previous 3D industrial datasets \cite{DBLP:conf/nips/LiuXCLWLWZ23,DBLP:conf/visapp/BergmannJSS22,DBLP:conf/accv/BonfiglioliTSFG22,Li_2024_CVPR} are mostly built based on toy, model, or synthetic data, which is not suitable for real industrial scenarios. To ensure the authenticity of the data, both samples of CPS3D-Seg are all from actual production lines. What's more, we collected 20 different types of products to ensure data diversity.

\begin{figure}[tp]
    \centering
    \includegraphics[width=0.8\linewidth]{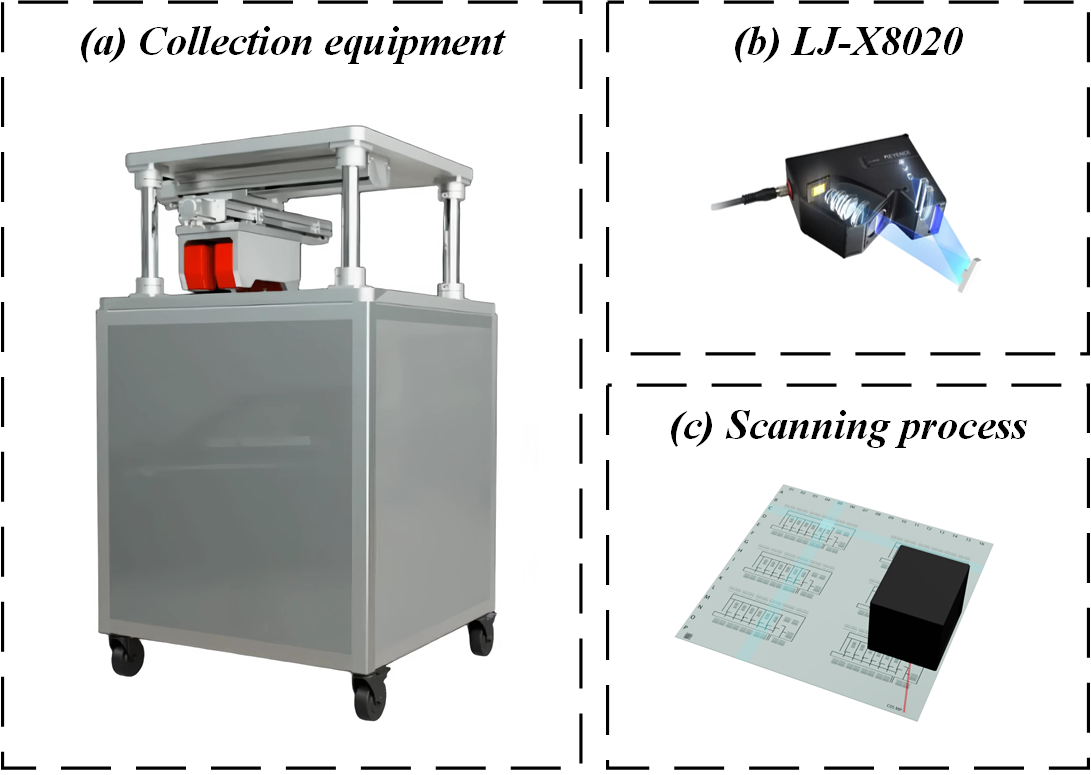}
    \caption{(a) The collection equipment to capture point cloud data. (b) The Keyence LJ-X8020 line laser scanner. (c) The biaxial motion mechanism controls the line laser, scanning from one corner of the sample and covering the full area. }
    \label{fig:2}
\end{figure}

\textbf{Collection Equipment.} We employ high-resolution collection equipment to acquire precise point cloud data, as illustrated in Fig. \ref{fig:2}. The collection equipment comprises four LJ-X8020s, a dual-axis displacement mechanism, and a movable placement platform. Four LJ-X8020s are fixed below the dual-axis displacement mechanism through sheet metal connectors, and the dual-axis displacement mechanism is fixed to the movable placement platform by four adjustable height brackets. What's more, the control system of the collection equipment is integrated into the movable placement platform. Four LJ-X8020s, driven by a dual-axis displacement mechanism, start scanning from a particular corner of the sample and move in a zigzag pattern along the platform until it covers all the planes of the sample. 

\begin{figure}[tp]
    \centering
    \includegraphics[width=\linewidth]{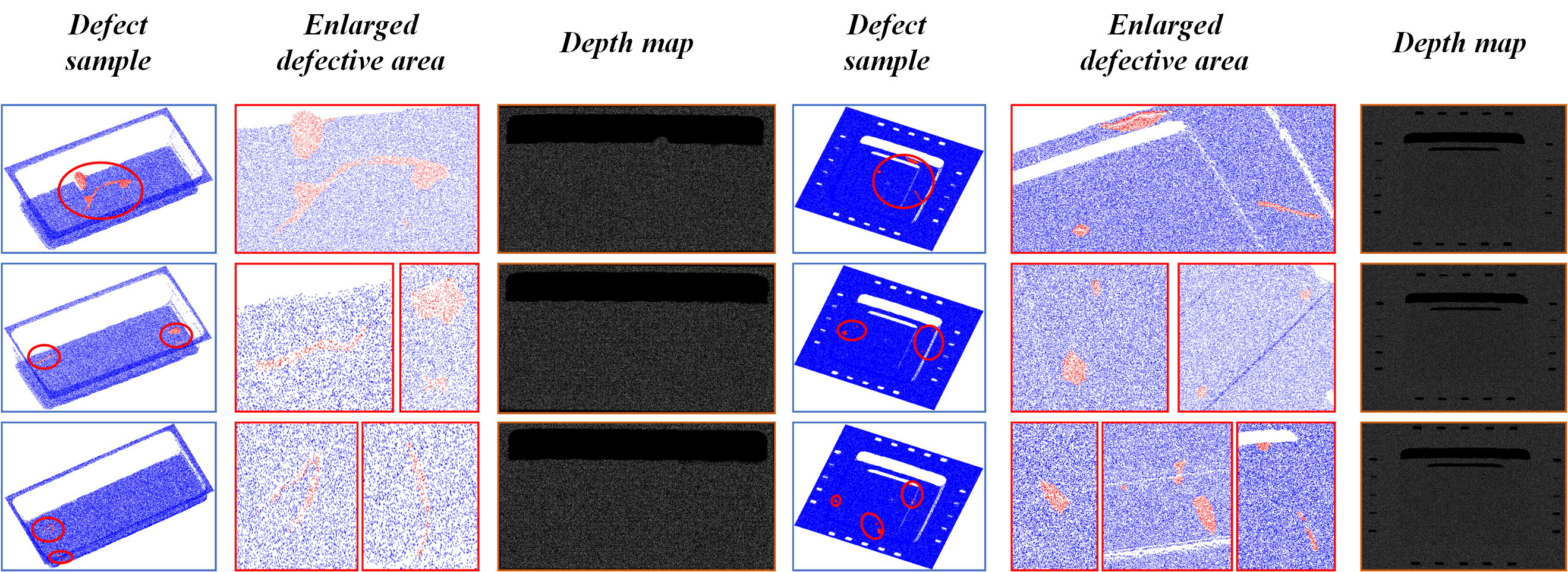}
    \caption{The concrete examples of CPS3D-Seg. }
    \label{fig:3}
\end{figure}

\textbf{Data Processing.} As each LJ-X8020 independently acquires partial point cloud data, we register the four laser heads using FRICP \cite{9336308}. Subsequently, we integrate the data from each LJ-X8020 according to the extrinsic parameters to provide a comprehensive sample point cloud. Our collection equipment automatically acquire high-precision data on a large scale in a brief amount of time, which presents significant challenges for data storage, transmission, and analysis. Consequently, we utilize 3D grids \cite{Lee2001PointDR} to eradicate many redundant data points in dense point clouds. 

\begin{table}[bp]
  \centering
  \resizebox{\linewidth}{!}{
    \begin{tabular}{c|cccccc}
    \toprule
    \toprule
    Dataset & Object Category & Sample Number & Point Resolution & Point Precision & Point Number & Type \\
    \midrule
    MVTecAD-3D & 10    & 3604 & 0.37 mm & 0.11 mm & 10K-20K & Real \\
    \rowcolor[rgb]{ .949,  .949,  .949} Eyecandies & 10    & \textbf{15500} & Not applicable & Not applicable & Not applicable & Syn \\
    Real3D-AD & 12    & 1200  & 0.04 mm & 0.011 mm & \textcolor[rgb]{ .188,  .329,  .588}{35K-780K} & Real \\
    \rowcolor[rgb]{ .949,  .949,  .949} Anomaly ShapeNet & \textbf{50} & 1600  & Not applicable & Not applicable & 8K-30K & Syn \\
    \midrule
    CPS3D-Seg (Ours) & 20 & 1300  & \textbf{0.0025 mm} & \textbf{0.0003 mm} & \textbf{114K-812K} & Real \\
    \bottomrule
    \bottomrule
    \end{tabular}}%
  \caption{The comparison between different datasets.}
  \label{tab:1}%
\end{table}%

\textbf{Annotation.} We pre-screen the defective areas of each sample using a high-magnification microscope and then annotate these areas in the corresponding point cloud data. We followed the commonly used annotation tool CloudCompare in point cloud segmentation task to implement this process. As shown in Fig. \ref{fig:3}, we display several defect samples and enlarge the defective area to present more details.

\begin{figure}[tp]
    \centering
    \includegraphics[width=\linewidth]{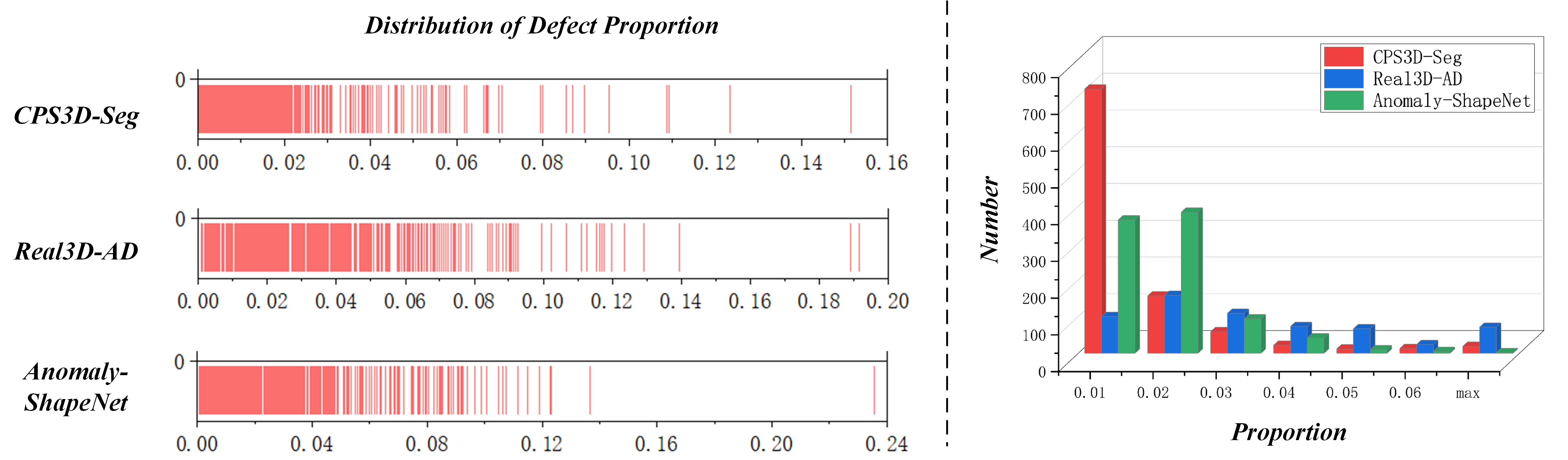}
    \caption{The defect proportion statistics of different datasets. CPS3D-Seg shows significant advantages in the defect proportion area, which makes our dataset more applicable and accurate in small defect detection. }
    \label{fig:c1}
\end{figure}

\textbf{Data Statistics \& Comparison with Other Datasets.} As shown in Tab. \ref{tab:1}, CPS3D-Seg consists of 1300 samples under 20 different product categories, and the number of point clouds in our dataset ranges from 114K to 812K, which is the best compared with other dataset. Considering the difficulty of data acquisition in the field of integrated circuits, our dataset still surpasses dataset Real3D-AD \cite{DBLP:conf/nips/LiuXCLWLWZ23}, which is also based on industrial parts. Compared with synthetic datasets (Eyecandies \cite{DBLP:conf/accv/BonfiglioliTSFG22} and Anomaly ShapeNet \cite{Li_2024_CVPR}), CPS3D-Seg exhibits limitations in both quantity and categories, but it offers more excellent practical research value due to the authenticity of the point cloud. In addition, thanks to the superiority of the device, CPS3D-Seg has the best resolution and precision. As shown in Fig. \ref{fig:c1}, we calculated the defect proportion of each sample in CPS3D-Seg, Real3D-AD, and Anomaly-ShapeNet. It can be clearly seen that the defect proportion in our dataset is mostly concentrated in the range of 0-0.01\%, which is a leading advantage compared to mainstream datasets, such as Real3D-AD and Anomaly-ShapeNet. This indicates that our data can provide scenarios for minor defect segmentation.

\textbf{Adaptation to other IC substrates.} Our data collection equipment is designed to be universal, ensuring consistent performance across diverse IC substrates. This capability means that whether the substrate is made from ceramic, resin, glass, or other materials, the equipment is able to capture the 3D data uniformly and reliably, without being affected by material properties. Moreover, 3D point cloud data primarily captures the geometric shape and structure of IC substrates, rather than material characteristics. This dataset reflects consistent shape and structural features of the substrates, independent of the material composition, supporting us in conducting research on various IC substrates.

\section{Methodology}
\label{sec:method}

\begin{figure}[tp]
    \centering
    \includegraphics[width=\linewidth]{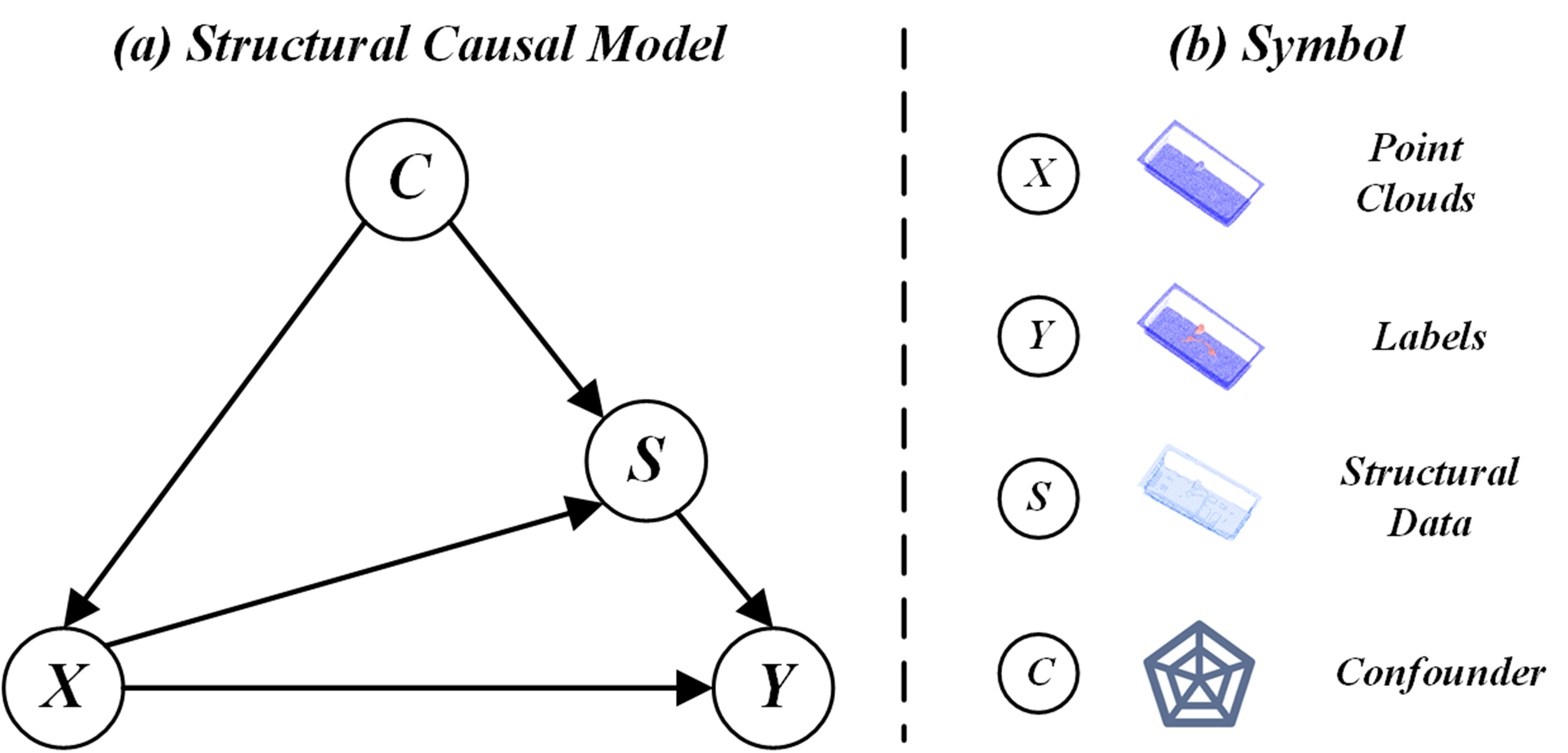}
    \caption{(a) The proposed structural causal model for point cloud. (b) The specific meaning of symbols.}
    \label{fig:4}
\end{figure}

\subsection{Preliminaries}
Recall in Fig. \ref{fig:1} that we illustrate the complexity confounder can affect the representation and structured processing of point cloud data. We formulate the causalities among point cloud $X$ and labels $Y$, with a Structural Causal Model (SCM). As illustrated in Fig. \ref{fig:4} (a) and (b), our SCM consists of confounder $C$, structural data $S$, point cloud $X$ and labels $Y$. All paths of SCM are as follow:

\textbf{Direct path} $X\rightarrow Y$. The point cloud $X$ directly affects the label $Y$. This path indicates that specific features can be directly derived from the point cloud and subsequently employed to predict labels. Intuitively, this can be interpreted as critical information regarding object boundaries and categories that may be incorporated into point cloud data. This information can be employed for 3D segmentation tasks without requiring intermediate steps. 

\textbf{Indirect path} $ X \rightarrow S \rightarrow Y$. The point cloud $X$ indirectly affects the labels $Y$ through structural data $S$. This path denotes the point cloud data being first extracted into a structural data representation $S$ using Octree, Voxel, or others, which is then used to predict labels $Y$. Compared to the direct path, the indirect path can extract high-level features from $S$, making it easier to predict $Y$.

\textbf{Confounder path} $X \leftarrow C \rightarrow S \rightarrow Y $. The confounder $C$ affects the point cloud $X$ and structural data $S$, affecting the labels $Y$. Here, the confounder factor $C$ represents the complexity of the point cloud, which may include its density, noise level, and other factors. $C$ affects the formation of point clouds $C \rightarrow X$, as well as the generated structural data $C \rightarrow S$, ultimately affecting the labels $Y$. Consequently, $C$ introduces a non-causal path that requires further adjustment to control this confounder.

\textbf{Causal intervention by backdoor adjustment.} Since $C$ is a confounder factor that affects $X$ and $S$, and $S$ is a mediating variable from $X$ to $Y$, we need to adjust $C$ to block the backdoor path, but we should not adjust $S$ because it is the mediator of $X$'s impact on $Y$. We utilize conditional probabilities to represent the causal effects between variables:

% As the intervention is not observable, we firtly cut off the path $ X \rightarrow S$ and stratify $C$ into piece $C={c}$. Formally, we have: 

% \begin{align}
%   P(Y|do(X))= \sum_{c} P(Y|X, S=f(X,c))P(c)
% \label{f1}
% \end{align}

% where $f(\cdot )$ is a function that computes the structural features corresponding to each point cloud $x$. Since $C$ is no longer connected with $X$, the causal intervention enables $X$ to equitably integrate each region into $Y$'s prediction, contingent upon a prior $P(c)$.

\begin{flalign}
P(Y|do(X=x))= \sum_{c} P(Y|X=x, C=c)P(C=c)
\label{f1}
\end{flalign}

where the $do$-calculus is the causal intervention in \cite{casualbook}, which help us eliminate the $S$-related terms on $X$. 

\begin{figure*}[tp]
    \centering
    \includegraphics[width=\linewidth]{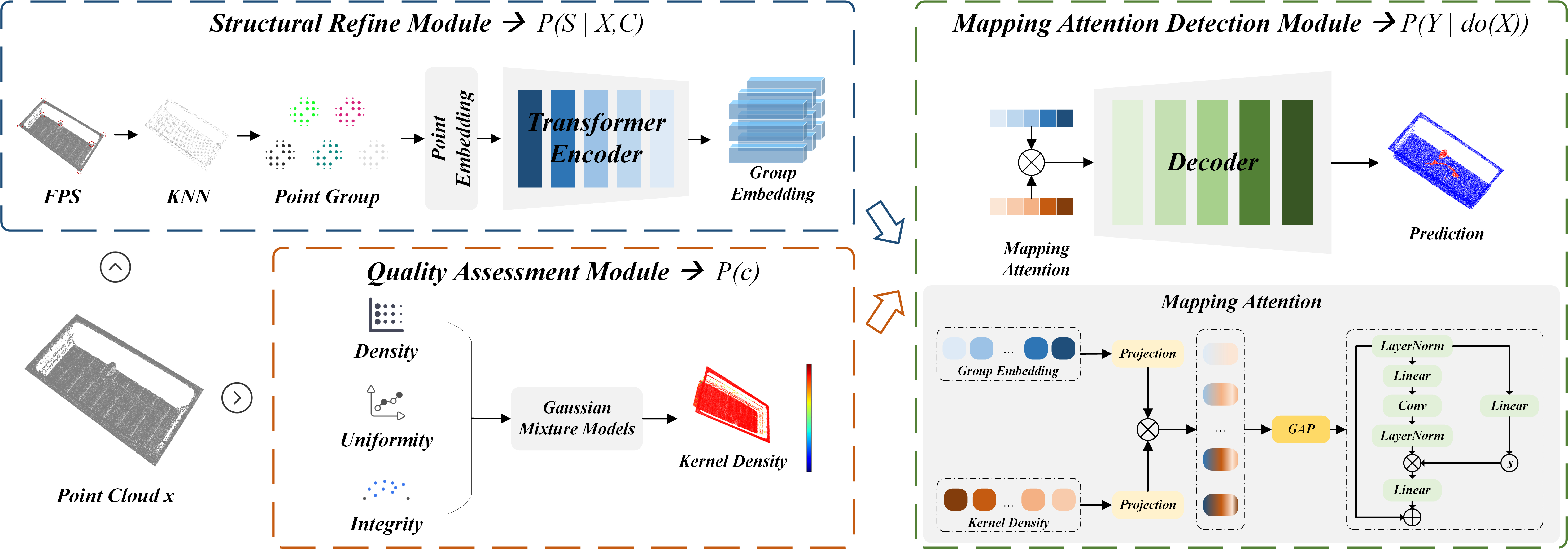}
    \caption{The proposed point cloud  via causal intervention.}
    \label{fig:5}
\end{figure*}

In order to achieve the backdoor adjustment process, we make the following assumption:

\textbf{Assumption 1:} \textit{When $S$ is a sufficient statistical measure of $X$, S contains all the information related to $Y$ in $X$. Given $S$, $Y$ is independent of $X$ ($Y \bot  X \mid S$).}

Therefore, we can derive the following theorem:

\textbf{Theorem 1:} \textit{Under the causal graph in Fig. \ref{fig:4}, suppose the real-world point cloud data $X$ can be represented by $S$. The causal effect is transformed into the following equation:}

\begin{flalign}
&P(Y|do(X=x)) =
\notag
\\ & \sum_{c} \sum_{s} P(Y|S=s)P(S=s|X=x, C=c)P(C=c)
\label{f4}
\end{flalign}

In this task, we use $P(Y|do(X=x))$ as the new point-level classifier. The complete proof for the theorem can be found in the supplementary material.

\subsection{The Proposed CINet}
To prevent the detection process from overfitting to specific hidden confounders, we propose a causal intervention network (CINet) according to the SCM in Fig. \ref{fig:4} (a). As indicated in Eq. \ref{f4}, $C$ is a significant factor, and its solving process will affect the final result. However, $C$ is unobservable in the point cloud segmentation task. In order to virtually evaluate $P(c)$, we conduct an overall rating mechanism including density, uniformity, and integrity of realistic point cloud to approximately replace potential impacts, $e.g. C={c_i}_1^{n}$, where $n$ is the data size in our dataset and $c_i$ represents to the kernel density of each point cloud $x$. Specifically, we develop three sub-modules, i.e., a quality assessment module, a structural refinement module, and a mapping attention detection module to maximize $P(Y|do(X=x))$ for point cloud . The details of CINet are illustrated in Fig. \ref{fig:5}.

\textbf{Quality assessment module.} For $P(c)$, we utilize the Gaussian Mixture Model (GMM) to capture different dimensions of measurement and information, $i.e.$, density reflects the degree of concentration of points, uniformity reflects the consistency of point distribution, and integrity may reveal the comprehensiveness of data. Individual eigenvalues may provide limited perspectives, but linking them as vectors can reveal more macroscopic attributes. 

Firstly, we use the kernel density estimation (KDE) method to calculate the density of points in space. Secondly, we divide the point cloud into a uniform grid and calculate the number of points in each grid. The smaller the difference in the number of points in each grid, the higher the uniformity. Thirdly, we measure the integrity of point clouds by calculating the continuity of normal directions and surface features. For each point cloud sample $x_i$, we combine the above features to acquire $\mathbf{c}_i = (c_{\text{density}, i}, c_{\text{uniformity}, i}, c_{\text{completeness}, i})$. The density function of GMM can be formulated as:

\begin{align}
P(\mathbf{c}) = \sum_{k=1}^{K} \pi_k \cdot \mathcal{N}(\mathbf{c} | \boldsymbol{\mu}_k, \boldsymbol{\Sigma}_k)
\label{f5}
\end{align}

where $K$ is the total number of Gaussian components in the GMM model. $\pi_k$ is the mixing coefficient of the $K-th$ Gaussian component, representing its contribution to the overall model, and satisfying $\sum_{k=1}^{K} \pi_k = 1$.

The specific form of Gaussian distribution in $Eq.$ \ref{f5} is as follows:
\begin{align}
\mathcal{N}(\mathbf{c} | \boldsymbol{\mu}_k, \boldsymbol{\Sigma}_k) = \frac{\exp\left(-\frac{1}{2} (\mathbf{c} - \boldsymbol{\mu}_k)^T \boldsymbol{\Sigma}_k^{-1} (\mathbf{c} - \boldsymbol{\mu}_k)\right)}{\sqrt{(2\pi)^3 |\boldsymbol{\Sigma}_k|}} 
\label{f6}
\end{align}

where $|\boldsymbol {\Sigma} _k |$ is the determinant of the covariance matrix, $(\mathbf{c} - \boldsymbol{\mu} _k) ^ T$ is the transpose of the deviation between the eigenvector and the mean, $\boldsymbol {\Sigma}_k^{-1}$ is the inverse of the covariance matrix.

Then, we use expectation maximization (EM) method for iterative solution. Firstly, we calculate the posterior probability of each data point for each Gaussian component through E-step:

\begin{align}
    \gamma_{ik} = \frac{\pi_k \cdot \mathcal{N}(\mathbf{c}_i | \boldsymbol{\mu}_k, \boldsymbol{\Sigma}k)}{\sum{j=1}^{K} \pi_j \cdot \mathcal{N}(\mathbf{c}_i | \boldsymbol{\mu}_j, \boldsymbol{\Sigma}j)}
\label{f7}
\end{align}

where $\gamma_{ik}$ is the sample $\mathbf{c}_i$ the probability of belonging to the $K-th$ component.

Then, we update the parameters of each Gaussian component through M-step:

\begin{align}
\boldsymbol{\mu}_k &= \frac{\sum_{i=1}^{N} \gamma_{ik} \mathbf{c}_i}{\sum_{i=1}^{N} \gamma_{ik}} \notag \\
\boldsymbol{\Sigma}_k &= \frac{\sum_{i=1}^{N} \gamma_{ik} (\mathbf{c}_i - \boldsymbol{\mu}_k)(\mathbf{c}_i - \boldsymbol{\mu}_k)^T}{\sum_{i=1}^{N} \gamma_{ik}}  \\
\pi_k &= \frac{\sum_{i=1}^{N} \gamma_{ik}}{N} \notag
\label{f8}
\end{align}

By repeatedly performing these two steps until the model converges, we can effectively replace the probability representation of $C$ with the observed data features. 

\textbf{Structural refinement module.} For $P(S|X, C)$, we aim to extract a group embedding to reorder the structure of each point cloud $x$. Specifically, we initially employ the farthest point sampling (FPS) to select $n$ keypoints. Then, we use the K-Nearest Neighbors (KNN) algorithm to aggregate point clouds around each keypoint, 

\begin{align}
    \mathcal{G}_i = { point \in x_i \ | \ \text{dist}(point, \mathbf{key}_j) \leq \text{Threshold}_K }
\end{align}

where $\text{dist}$ represents the distance between each point and key point, $\text {Threshold}_K$ is the maximum neighborhood distance threshold defined with K as the parameter. We can acquire point group $\mathcal{G}_i = \left \{ g_1, \dots, g_n \right \}$. 

For each point group, we project the position into a standard embedding layer of Transformer:

\begin{align}
    \mathbf{e}_j = \text{Linear}(\mathbf{p}_j) = \mathbf{W} \cdot \mathbf{p}_j + \mathbf{b}
\label{f10}
\end{align}

The sums of the above point embeddings are fed into the encoder to generate group embeddings $E_g$. The Transformer's remarkable capacity helps us to model long-range connections to capture intra-group dependencies. 

\textbf{Mapping attention detection module.} Following the pipeline of Eq. \ref{f4}, we propose a mapping attention mechanism to integrate the group embedding and the kernel density. Specifically, given the group embedding and kernel density, we employ a $1 \times 1$ projection layer to map the features into the same latent space. Then, we combine them via matrix multiplication to acquire a weighted group embedding. Due to the limited number of point clouds within different groups, it is not easy to characterize the overall characteristics. Thus, we utilize global average pooling (GAP) to integrate global information. Compared to fully connected layers, GAP eliminates extensive connections and numerous parameter computations while maintaining the significance of various spatial positions in group embeddings. Following the acquisition of global features, we propose a block including a primary path and a residual path to enhance global representation, thereby mitigating feature redundancy and overfitting to provide compact and expressive representations.

\section{Experiments}
\label{sec:experiment}

\subsection{Benchmark \& Comparison Construction } 
We have built a detailed and comprehensive benchmark to evaluate the effectiveness of CPS3D-Seg dataset. This benchmark covers classic and latest point cloud segmentation algorithms from four different categories, such as CNN-based methods \cite{8099499, 8953494,10.5555/3295222.3295263,Horwitz_2023_CVPR,wang2023multimodal,9879738,DBLP:conf/nips/LiuXCLWLWZ23,qian2022pointnext,10204778,peng2024oacnns}, graph-based methods \cite{dgcnn,Wang2019_GACNet, 9156514,lei2020spherical,10.1109/TPAMI.2023.3238516,robert2024scalable}, transformer-based methods \cite{DBLP:conf/iccv/ZhaoJJTK21,DBLP:conf/nips/0002LJLZ22,wu2024ptv3,10543016,Wang2023OctFormer,kolodiazhnyi2024oneformer3d}, and mamba-based methods \cite{zhang2024point,han2024mamba3d,liu2024point,liang2024pointmamba,Wang2024PoinTrambaAH}. Our goal is to provide a fair and systematic performance analysis on CPS3D-Seg dataset. To ensure fairness in comparison, all models were trained and tested in the same computing environment. During the training process, we used the same ratio of data to ensure comparability of the results. 

\begin{table}[htbp]
  \centering
 
    \resizebox{\linewidth}{!}{
    \begin{tabular}{c|c|ccc|cc|cc}
    \toprule
    \toprule
    \multirow{2}[4]{*}{Methods} & \multirow{2}[4]{*}{Publication \& Year} & \multicolumn{3}{c|}{Average Metrics} & \multicolumn{2}{c|}{Normal Metrics} & \multicolumn{2}{c}{Abnormal Metrics} \\
\cmidrule{3-9}          &       & mIoU  & mAP   & OA    & iou   & Precision & iou   & Precision \\
    \midrule
    \multicolumn{9}{c}{CNN-Based} \\
    \midrule
    \rowcolor[rgb]{ .949,  .949,  .949} Pointnet & CVPR 2017 & 0.6342 & 0.6548 & 0.9896 & 0.9900 & 0.9980 & 0.2790 & 0.3110 \\
    Pointnet++ & NeurIPS 2017 & 0.7853 & 0.8556 & 0.9933 & 0.9930 & 0.9970 & 0.5770 & 0.7140 \\
    \rowcolor[rgb]{ .949,  .949,  .949} SPUNet & CVPR 2019 & 0.8093 & 0.8502 & 0.9946 & 0.9945 & 0.9984 & 0.6242 & 0.7020 \\
    Pointnext & NeurIPS 2022 & 0.7794 & 0.8030 & 0.9940 & 0.9939 & 0.9990 & 0.5648 & 0.6070 \\
    \rowcolor[rgb]{ .949,  .949,  .949} Pointvector & CVPR 2023 & 0.7836 & 0.8124 & 0.9940 & 0.9939 & 0.9988 & 0.5732 & 0.6260 \\
    KPConvx & CVPR 2024 & 0.7970 & 0.8502 & 0.9940 & 0.9939 & 0.9978 & 0.6000 & 0.7026 \\
    \rowcolor[rgb]{ .949,  .949,  .949} OACNNs & CVPR 2024 & 0.7991 & 0.8380 & 0.9943 & 0.9942 & 0.9984 & 0.6039 & 0.6775 \\
    \midrule
    \multicolumn{9}{c}{Graph-Based} \\
    \midrule
    \rowcolor[rgb]{ .949,  .949,  .949} DGCNN & ACM TOG 2019 & 0.5547 & 0.5607 & 0.9886 & 0.9886 & 0.9990 & 0.1208 & 0.1224 \\
    GACNet & CVPR 2019 & 0.7724 & 0.8065 & 0.9914 & 0.9910 & 0.9940 & 0.5530 & 0.6190 \\
    \rowcolor[rgb]{ .949,  .949,  .949} 3D-GCN & CVPR 2020 & 0.7802 & 0.8406 & 0.9812 & 0.9891 & 0.9921 & 0.5712 & 0.6891 \\
    SPH-3D GCN  & TPAMI 2020 & 0.7861 & 0.7912 & 0.9891 & 0.9902 & 0.9911 & 0.5819 & 0.5912 \\
    \rowcolor[rgb]{ .949,  .949,  .949} AGConv & TPAMI 2023 & 0.7837 & 0.8084 & 0.9902 & 0.9879 & 0.9877 & 0.5795 & 0.6291 \\
    SGC   & 3DV 2024 & 0.7995 & 0.8395 & 0.9945 & 0.9899 & 0.9889 & 0.6091 & 0.6901 \\
    \midrule
    \multicolumn{9}{c}{Transformer-Based} \\
    \midrule
    \rowcolor[rgb]{ .949,  .949,  .949} PTV1  & ICCV 2021 & 0.8009 & 0.8549 & 0.9941 & 0.9940 & 0.9978 & 0.6078 & 0.7120 \\
    PTV2  & NeurIPS 2022 & 0.8182 & 0.8613 & 0.9948 & \textcolor[rgb]{ 0,  .439,  .753}{\textbf{0.9947}} & 0.9983 & 0.6417 & 0.7244 \\
    \rowcolor[rgb]{ .949,  .949,  .949} Octformer & ACM TOG 2023 & 0.7893 & 0.8129 & 0.9943 & 0.9942 & 0.9991 & 0.5843 & 0.6268 \\
    SPT   & ICCV  2023 & 0.6899 & 0.7790 & 0.9886 & 0.9886 & 0.9942 & 0.3900 & 0.5638 \\
    \rowcolor[rgb]{ .949,  .949,  .949} PointNAT & TGRS 2024 & 0.7973 & 0.8190 & 0.9913 & 0.9934 & 0.9902 & 0.6012 & 0.6477 \\
    OneFormer3D & CVPR 2024 & 0.8118 & 0.8449 & 0.9933 & 0.9893 & 0.9899 & 0.6343 & 0.6998 \\
    \rowcolor[rgb]{ .949,  .949,  .949} LPFP  & TIP  2024 & 0.7712 & 0.8252 & 0.9926 & 0.9927 & 0.9912 & 0.5497 & 0.6591 \\
    GSTran & ICPR 2024 & 0.7847 & 0.8272 & 0.9919 & 0.9929 & 0.9917 & 0.5764 & 0.6626 \\
    \rowcolor[rgb]{ .949,  .949,  .949} PDF   & CVPR 2024 & 0.7950 & 0.8379 & 0.9938 & 0.9931 & 0.9945 & 0.5968 & 0.6812 \\
    SPG   & ECCV 2024 & 0.7969 & 0.8327 & 0.9954 & 0.9933 & 0.9925 & 0.6005 & 0.6728 \\
    \rowcolor[rgb]{ .949,  .949,  .949} PTV3  & CVPR 2024 & \textcolor[rgb]{ 0,  .439,  .753}{\textbf{0.8204}} & \textcolor[rgb]{ 0,  .439,  .753}{\textbf{0.8682}} & 0.9948 & \textcolor[rgb]{ 0,  .439,  .753}{\textbf{0.9947}} & 0.9981 & \textcolor[rgb]{ 0,  .439,  .753}{\textbf{0.6461}} & \textcolor[rgb]{ 0,  .439,  .753}{\textbf{0.7384}} \\
    \midrule
    \multicolumn{9}{c}{Mamba-Based} \\
    \midrule
    \rowcolor[rgb]{ .949,  .949,  .949} PCM   & Arxiv 2024 & 0.7472 & 0.7734 & 0.9963 & 0.9963 & \textcolor[rgb]{ 0,  .439,  .753}{\textbf{0.9985}} & 0.4980 & 0.5641 \\
    Point-Mamba & Arxiv 2024 & 0.6529 & 0.6881 & 0.9873 & 0.9872 & 0.9970 & 0.3186 & 0.3792 \\
    \rowcolor[rgb]{ .949,  .949,  .949} Mamba3D & ACM MM 2024 & 0.7271 & 0.7703 & 0.9891 & 0.9871 & 0.9914 & 0.4671 & 0.5491 \\
    PointMamba & Arxiv 2024 & 0.7494 & 0.7972 & 0.9912 & 0.9899 & 0.9931 & 0.5089 & 0.6012 \\
    \rowcolor[rgb]{ .949,  .949,  .949} PoinTramba & Arxiv 2024 & 0.7377 & 0.7920 & 0.9931 & 0.9902 & 0.9938 & 0.4851 & 0.5901 \\
    \midrule
    Ours  & -     & \textcolor[rgb]{ 1,  0,  0}{\textbf{0.8544}} & \textcolor[rgb]{ 1,  0,  0}{\textbf{0.8997}} & \textcolor[rgb]{ 1,  0,  0}{\textbf{0.9953}} & \textcolor[rgb]{ 1,  0,  0}{\textbf{0.9989}} & \textcolor[rgb]{ 1,  0,  0}{\textbf{0.9996}} & \textcolor[rgb]{ 1,  0,  0}{\textbf{0.7098}} & \textcolor[rgb]{ 1,  0,  0}{\textbf{0.7998}} \\
    \bottomrule
    \bottomrule
    \end{tabular}}%
     \caption{The comparison results of different methods.}
  \label{tab2}%
\end{table}%

\textbf{Quantitative results.} The detail of our benchmark is shown in Tab. \ref{tab2}. The bold red font indicates the best result, while the bold blue font represents the second-best result. In the CNN-based method, SPUNet performs the best with a mIoU of 0.8093 and performs well on normal and abnormal metrics. In the graph-based method, SGC performs well with a mIoU of 0.7995, and the mIoU of abnormal indicators reaches 0.6091, which is relatively outstanding. PTV3 performs best in transformer-based methods, with a mIoU of 0.8204. At the same time, the mIoU and Acc on the abnormal indicators are also relatively high, at 0.6461 and 0.7384, respectively. In the mamba-based method, PointMamba performs the best, with a mIoU of 0.7494 and an abnormal indicator mIoU of 0.5089. Compared to the above methods, CINet achieved the best results in all metrics. Specifically, CINet outperforms PTV3 by approximately 4.16\% on the most important mIoU metric.

\begin{figure}[tp]
    \centering
    \includegraphics[width=\linewidth]{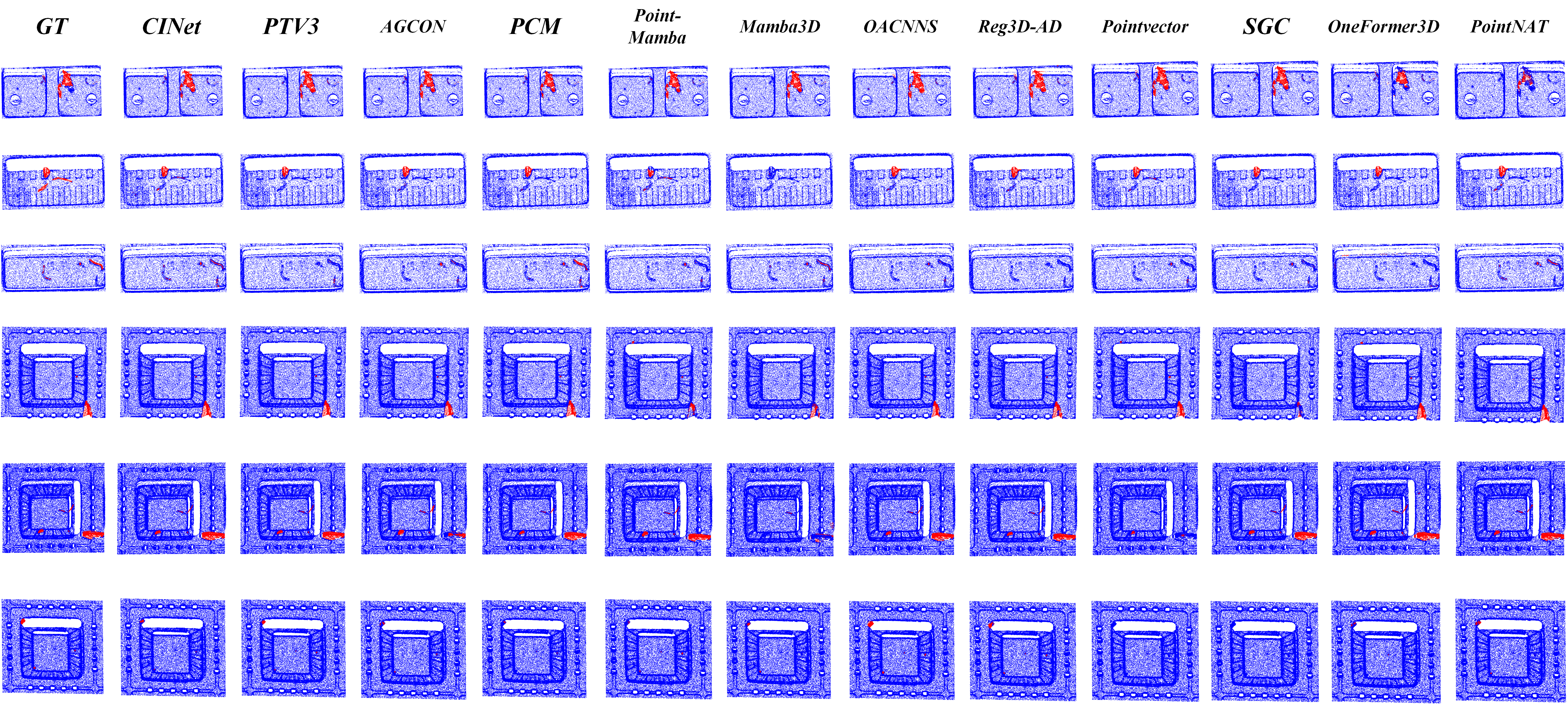}
    \caption{The visualization maps between different methods.}
    \label{fig:6}
\end{figure}

\textbf{Qualitative results.} To elucidate the contrast among various methods, we visualized the predicted maps of each model, as illustrated in Fig. \ref{fig:6}. It can be clearly seen from the qualitative results that there are differences in segmentation details and edge processing among various models. Our approach excels in segmenting the target edge, exhibiting distinct segmentation contours that proficiently differentiate various objects. Simultaneously, it exhibits enhanced resilience in managing intricate backdrops and attains a more equitable segmentation outcome.

% Through qualitative and quantitative research, our benchmark reveals that several models exhibit distinct benefits across multiple categories and metrics, offering a novel perspective for a more profound comprehension of point cloud segmentation tasks.

\begin{table}[htbp]
  \centering
    \resizebox{\linewidth}{!}{
    \begin{tabular}{ccc|ccc}
    \toprule
    \toprule
    \multicolumn{3}{c|}{Model Variants} & \multicolumn{3}{c}{Average Metrics} \\
    \midrule
    SR    & QA    & MAD   & mIoU  & mAcc  & allAcc \\
    \midrule
    \rowcolor[rgb]{ .949,  .949,  .949}       &       &       & 0.6832 & 0.7011 & 0.8652 \\
    \checkmark (\textbf{Q1})     &       &       & 0.7576(↑10.89\%) & 0.8021(↑14.40\%) & 0.9031(↑4.38\%) \\
    \rowcolor[rgb]{ .949,  .949,  .949}       & \checkmark (\textbf{Q2})     &       & 0.7682(↑12.44\%) & 0.8132(↑15.99\%) & 0.9213(↑6.48\%) \\
          &       & \checkmark (\textbf{Q3})    & 0.7891(↑15.50\%) & 0.8321(↑18.68\%) & 0.9321(↑7.73\%) \\
    \rowcolor[rgb]{ .949,  .949,  .949} \checkmark     & \checkmark     & \checkmark     & 0.8544(↑25.06\%) & 0.8997(↑28.33\%) & 0.9953(↑15.04\%) \\
    \bottomrule
    \bottomrule
    \end{tabular}}%
    \caption{The experiment results of ablation study.}
  \label{tab:3}%
\end{table}%

\begin{figure}[tp]
    \centering
    \includegraphics[width=\linewidth]{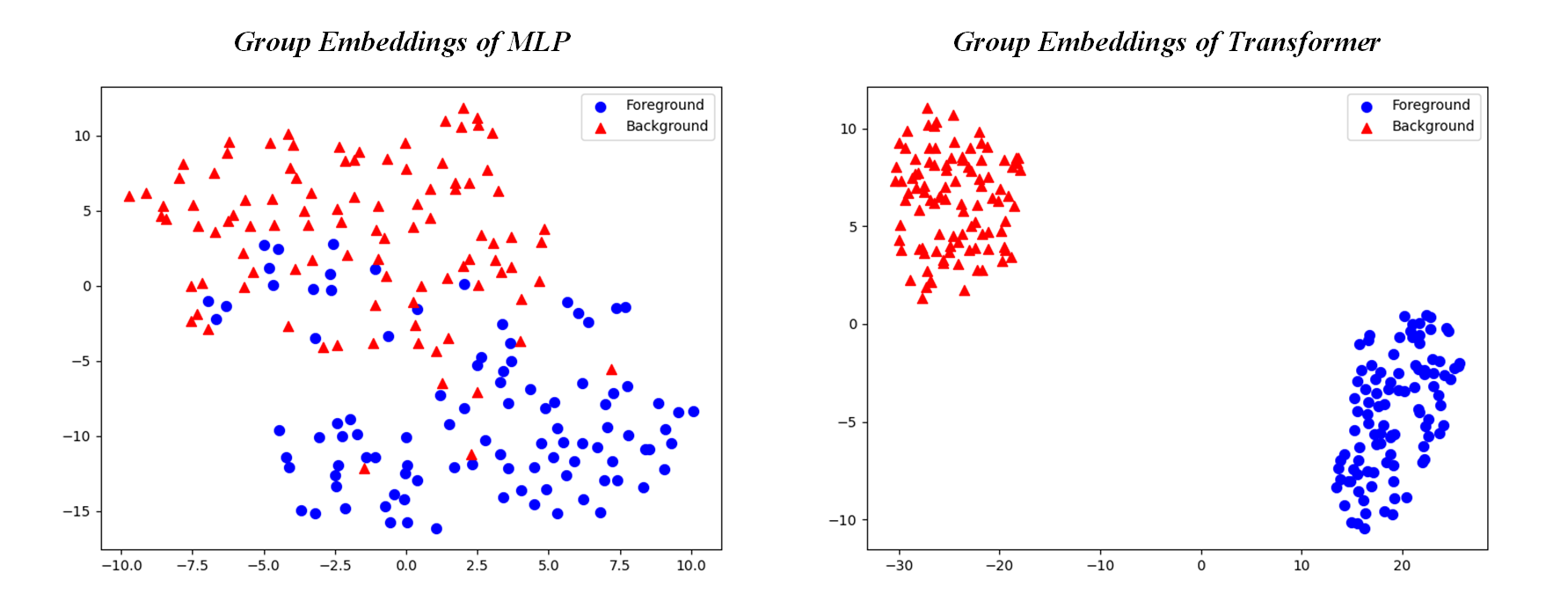}
    \caption{The T-SNE visualization.}
    \label{fig:7}
\end{figure}

\subsection{Ablation Study}
We conducted ablation studies to answer the following questions. \textbf{Q1:} \textit{Can structural refine module effectively capture structural information of ceramic package substrate?} \textbf{Q2:} \textit{Can the evaluation metrics selected by quality assessment module compensate for the inherent problems in the point cloud collection process?} \textbf{Q3:} \textit{Can attention detection module improve the characterization ability of defect features?}  

We design a baseline of CINet to evaluate the impact of each module. Specifically, we use MLP to generate group embeddings instead of a transformer encoder. We integrate different quality metrics as a vector rather than using GMM to calculate kernel density. Moreover, we do not use mapping attention mechanisms but directly concatenate the group embeddings and quality features.

\textbf{A1:} Results in Tab. \ref{tab:3} (\textbf{Q1}) show that the SR module improve the mIoU ($\uparrow 10.89\%$), mAcc ($\uparrow 14.40\%$), and allAcc ($\uparrow 4.38\%$). Moreover, we use T-SNE to visualize the different group embeddings extracted by MLP and transformer encoder. As shown in Fig. \ref{fig:7}, transformer-based SR modules present clearer structural boundaries between foreground and background.

\begin{figure}[tp]
    \centering
    \includegraphics[width=\linewidth]{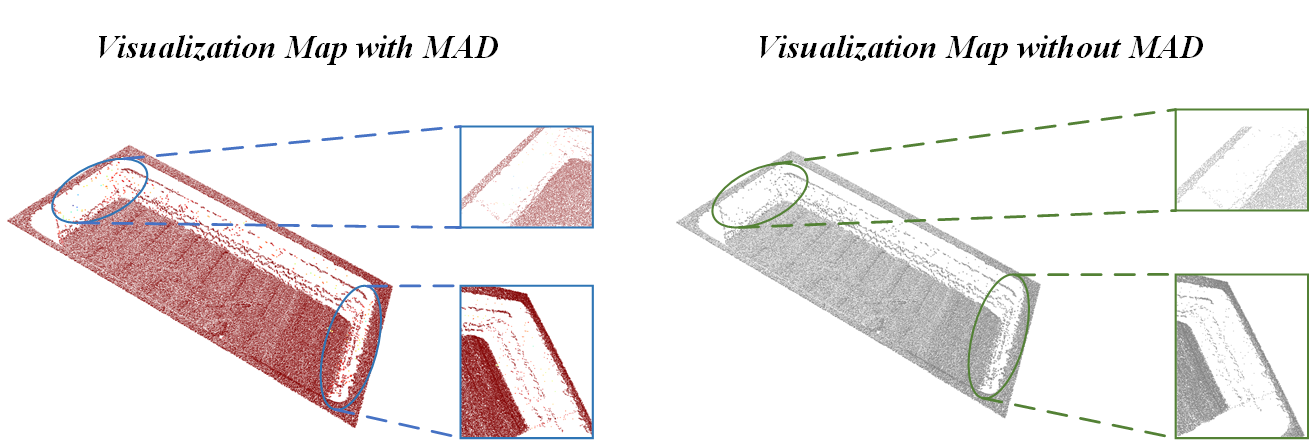}
    \caption{The visualization maps with and without MAD.}
    \label{fig:8}
\end{figure}

\begin{table}[htbp]
  \centering

  \resizebox{\linewidth}{!}{
    \begin{tabular}{ccc|ccc}
    \toprule
    \toprule
    \multicolumn{3}{c|}{Metric} & \multicolumn{3}{c}{Average Metrics} \\
    \midrule
    Density & Uniformity & Integrity & mIoU  & mAcc  & allAcc \\
    
    \rowcolor[rgb]{ .949,  .949,  .949} \checkmark     &       &       & 0.8012 & 0.8381 & 0.9782 \\
          & \checkmark     &       & 0.8098 & 0.8401 & 0.9821 \\
    \rowcolor[rgb]{ .949,  .949,  .949}       &       & \checkmark     & 0.8021 & 0.8397 & 0.9796 \\
    \checkmark     & \checkmark     &       & 0.8132 & 0.8631 & 0.9925 \\
    \rowcolor[rgb]{ .949,  .949,  .949} \checkmark    &       & \checkmark     & 0.8201 & 0.8696 & 0.9912 \\
          & \checkmark     & \checkmark     & 0.8231 & 0.8622 & 0.9914 \\
    \rowcolor[rgb]{ .949,  .949,  .949} \checkmark     & \checkmark     & \checkmark     & 0.8544 & 0.8997 & 0.9953 \\
    \bottomrule
    \bottomrule
    \end{tabular}}%
      \caption{The results of different metrics used in QA module.}
  \label{tab:4}%
\end{table}%

\textbf{A2:} From Tab. \ref{tab:3} (\textbf{Q2}), we can observe that using the QA module can boost the performance when compared with the baseline. Moreover, we also evaluate the effectiveness of each metric used in the QA module, as shown in Tab. \ref{tab:4}.

\textbf{A3:} Results in Tab. \ref{tab:3} (\textbf{Q3}) indicate that the MAD module brings the most significant performance improvement than other modules. Moreover, we propose visualization maps with and without the MAD module. As shown in Fig. \ref{fig:8}, MAD provides weighted information for each point cloud. The process of changing from blue to red indicates an increasing importance in point cloud segmentation and it can provide additional information for subsequent predication. 

\section{Conclusion}
\label{sec:conclusion}

In this paper, we propose a novel CPS3D-Seg dataset for ceramic package substrates 3D  task. The high-precision data acquisition system composed of four line laser scanners ensures the quality of collected point clouds, far exceeding existing 3D  datasets. We conduct a comprehensive benchmark of CPS3D-Seg, which evaluates the performance of the dataset. What's more, we propose a point cloud  model using causal intervention (CINet). By constructing a structural causal model (SCM), we indirectly quantify the evaluation of confounding factors, which aims to eliminate the potential effect of the missing data and the aggregation errors. Extensive experiments proves the effectiveness of CINet. 

\section{Acknowledgment}
This work is supported by the National Key Research and Development Program of China (Grant 2024YFB3409202) and the Key Research and Development Program of Liaoning Province (Grant 2024020969-JH2/1024).

\bibliography{aaai2026}
\clearpage
\appendix

\begin{appendices}

\setcounter{table}{0}   %从0开始编号，显示出来表会A1开始编号
\setcounter{figure}{0}

\section{Appendix A: Proof of the Theorem 1}
\subsection{A.1. The Construction of Structural Causal Model}
As indicated in Fig. \ref{fig:4}, we construct a structural causal model for point cloud segmentation. The details of variable definitions and function relationships of SCM are as follows:

\[
\scriptsize
\begin{aligned}
&\bullet\ \text{Confounder} & C &:= (\rho, \sigma, \iota) \in \mathbb{R}^3 \quad (\text{Density/Uniformity/Completeness}) \\
&\bullet\ \text{Point Clouds} & X &:= f(C) + \epsilon_X \quad \epsilon_X \sim \mathcal{N}(0,\Sigma_X) \\
&\bullet\ \text{Structural Data} & S &:= g(X, C) + \epsilon_S \quad \epsilon_S \sim \mathcal{N}(0,\Sigma_S) \\
&\bullet\ \text{Labels} & Y &:= h(X) \oplus k(S) + \epsilon_Y \quad \epsilon_Y \sim \text{Bern}(p)
\end{aligned}
\]

The paths in SCM are divided into direct path, indirect path, and confounder path. We utilize conditional probabilities to represent the causal effects between variables:

\[
\begin{aligned}
    P(Y|do(X=x)) = \sum_{c} P(Y|X=x, C=c)P(C=c) \quad \text{(1)}
\end{aligned}
\]

According to SCM, $Y$ is determined by $S$ and $X$ together:

\[
\begin{aligned}
    & P(Y|X=x, C=c) \\
    & = \sum_{s} P(Y|S=s, X=x, C=c)P(S=s|X=x, C=c)
\end{aligned}
\]

Substituting the above formula into (1) yields:

\[
\tiny
\begin{aligned}
    & P(Y|do(X=x)) \\
    & = \sum_{c} \sum_{s} \underbrace{P(Y|S=s, X=x, C=c)}_{\text{prediction}} \underbrace{P(S=s|X=x, C=c)}_{\text{structural generation}} \underbrace{P(C=c)}_{\text{confounder distribution}} \quad \text{(2)}
\end{aligned}
\]

According to our \textbf{Assumption 1}, given $S, Y$ is independent of $X$ and $C$ when $S$ contains all the predicted information of $Y$ in $X$:

\[
\begin{aligned}
P(Y|S=s, X=x, C=c) = P(Y|S=s) \quad \forall x,c
\end{aligned}
\]

Simplify formula (2) from \textbf{Assumption 1} to:

\[
\begin{aligned}
& P(Y|do(X=x)) = \\
& \sum_{c} \sum_{s} P(Y|S=s) P(S=s|X=x, C=c) P(C=c) \quad \text{(3)}
\end{aligned}
\]

\subsection{A.2. Implement Details}

Decompose formula (3) into learnable modules:

\[
\scriptsize
\boxed{
\begin{aligned}
P(Y|do(X=x)) = \mathbb{E}_{c \sim P(C)} \Bigg[ \mathbb{E}_{s \sim P(S|X=x,C=c)} \bigg[ \underbrace{P(Y|S=s)}_{\text{MAD}} \bigg] \Bigg]
\end{aligned}
}
\]

Specifically, QA module estimates $P (C=c) $:

\[
\begin{aligned}
P(C=c) = \text{GMM}(c; \pi_k, \mu_k, \Sigma_k) \quad \text{(4)}
\end{aligned}
\]

SR module generates $S | X, C $:

\[
\begin{aligned}
S = \text{Transformer}(\text{FPS}(X), \text{KNN}(X); \theta_{SR}) \quad \text{(5)}
\end{aligned}
\]

MAD module calculates $P (Y | S) $:

\[
\begin{aligned}
P(Y|S) = \sigma\Big(\text{GAP}(\text{Att}(S, \rho(c))); \theta_{MAD}\Big) \quad \text{(6)}
\end{aligned}
\]

\section{Appendix B: More Details of Experiments}
\subsection{B.1. Experimental Settings}
\textbf{Implementation details.} For the proposed CINet, the initial learning rate is set to 0.01, the epoch is set to 50, and the batch size is set to 32, respectively. Moreover, we select ADAM \cite{DBLP:journals/corr/KingmaB14} as optimizer to optimize CINet and use Kaiming initialization \cite{10.1109/ICCV.2015.123} to initialize the parameters of CINet. Both codes are implemented by Python 3.9 and PyTorch-1.10, and the environment is built on Rocky Linux 8. Furthermore, all the experiments are built on 8 Nvidia GeForce RTX 3090 GPUs.

\textbf{Evaluation metric.} For 3D point cloud segmentation task, the most common performance criteria include overall accuracy (OA), mean average precision (mAP), and mean intersection over union (mIoU). We follow these metrics to better evaluate the performance of different models.

\subsection{B.2. Performance on Other Datasets}
In order to further verify the effectiveness of our algorithm, we also evaluate its performance on other datasets, such as Real3D-AD and Anomaly-ShapeNet. It is worth noting that Real3D-AD and Anomaly-ShapeNet was applied to 3D anomaly detection task, and we discarded normal samples and only used abnormal samples to validate the segmentation algorithms. The result of Real3D-AD is shown in Tab. \ref{tab:5} and the result of Anomaly-ShapeNet is shown in Tab. \ref{tab:r1}.

\begin{table}[htbp]
  \centering
  \resizebox{\linewidth}{!}{
\begin{tabular}{c|c|ccc|cc|cc}
    \toprule
    \toprule
    \multirow{2}[4]{*}{Methods} & \multirow{2}[4]{*}{Publication \& Year} & \multicolumn{3}{c|}{Average Metrics} & \multicolumn{2}{c|}{Normal Metrics} & \multicolumn{2}{c}{Abnormal Metrics} \\
\cmidrule{3-9}          &       & mIoU  & mAP   & OA    & IoU   & AP    & IoU   & AP \\
    \midrule
    \rowcolor[rgb]{ .949,  .949,  .949} KPConvx & CVPR 2024 & 0.7846 & 0.8166 & 0.9942 & 0.9942 & 0.9987 & 0.5750 & 0.6344 \\
    OACNNs & CVPR 2024 & 0.8068 & 0.8352 & 0.9949 & 0.9948 & 0.9989 & 0.6187 & 0.6714 \\
    \rowcolor[rgb]{ .949,  .949,  .949} PTV3  & CVPR 2024 & \textcolor[rgb]{ 0,  .439,  .753}{\textbf{0.8710}} & 0.8961 & \textcolor[rgb]{ 0,  .439,  .753}{\textbf{0.9967}} & 0.9966 & \textcolor[rgb]{ 1,  0,  0}{\textbf{0.9992}} & 0.7454 & 0.7930 \\
    SGC   & 3DV 2024 & 0.8756 & 0.9053 & 0.9967 & \textcolor[rgb]{ 0,  .439,  .753}{\textbf{0.9967}} & \textcolor[rgb]{ 0,  .439,  .753}{\textbf{0.9991}} & \textcolor[rgb]{ 0,  .439,  .753}{\textbf{0.7544}} & 0.8115 \\
    \rowcolor[rgb]{ .949,  .949,  .949} SPG   & ECCV 2024 & 0.8431 & 0.8941 & 0.9967 & 0.9956 & 0.9982 & 0.6906 & 0.7900 \\
    PCM   & Arxiv 2024 & 0.8498 & 0.8896 & 0.9959 & 0.9959 & 0.9986 & 0.7037 & 0.7805 \\
    \rowcolor[rgb]{ .949,  .949,  .949} PointMamba & Arxiv 2024 & 0.8586 & \textcolor[rgb]{ 0,  .439,  .753}{\textbf{0.9079}} & 0.9961 & 0.9961 & 0.9983 & 0.7212 & \textcolor[rgb]{ 1,  0,  0}{\textbf{0.8174}} \\
    Ours  & -     & \textcolor[rgb]{ 1,  0,  0}{\textbf{0.8823}} & \textcolor[rgb]{ 1,  0,  0}{\textbf{0.9081}} & \textcolor[rgb]{ 1,  0,  0}{\textbf{0.9969}} & \textcolor[rgb]{ 1,  0,  0}{\textbf{0.9969}} & \textcolor[rgb]{ 1,  0,  0}{\textbf{0.9992}} & \textcolor[rgb]{ 1,  0,  0}{\textbf{0.7678}} & \textcolor[rgb]{ 0,  .439,  .753}{\textbf{0.8170}} \\
    \bottomrule
    \bottomrule
    \end{tabular}}%
      \caption{The results on Real3D-AD dataset.}
  \label{tab:5}%
\end{table}%

\begin{table}[htbp]
  \centering
  \resizebox{\linewidth}{!}{
    \begin{tabular}{c|c|ccc|cc|cc}
    \toprule
    \toprule
    \multirow{2}[4]{*}{Methods} & \multirow{2}[4]{*}{Publication \& Year} & \multicolumn{3}{c|}{Average Metrics} & \multicolumn{2}{c|}{Normal Metrics} & \multicolumn{2}{c}{Abnormal Metrics} \\
\cmidrule{3-9}          &       & mIoU  & mAP   & OA    & IoU   & AP    & IoU   & AP \\
    \midrule
    \rowcolor[rgb]{ .949,  .949,  .949} KPConvx & CVPR 2024 & 0.6514 & 0.6666 & 0.9342 & \textcolor[rgb]{ 0,  .439,  .753}{\textbf{0.9242}} & 0.8987 & 0.3785 & 0.4344 \\
    OACNNs & CVPR 2024 & 0.7168 & 0.6752 & 0.8949 & 0.8948 & 0.7989 & 0.5387 & 0.5514 \\
    \rowcolor[rgb]{ .949,  .949,  .949} PTV3  & CVPR 2024 & 0.7460 & 0.7462 & 0.9067 & 0.9066 & \textcolor[rgb]{ 0,  .439,  .753}{\textbf{0.8992}} & 0.5854 & 0.5931 \\
    SGC   & 3DV 2024 & 0.7071 & 0.7053 & 0.9017 & 0.9097 & 0.8991 & 0.5044 & 0.5115 \\
    \rowcolor[rgb]{ .949,  .949,  .949} SPG   & ECCV 2024 & 0.6931 & 0.6448 & 0.8967 & 0.8956 & 0.7982 & 0.4906 & 0.4913 \\
    PCM   & Arxiv 2024 & 0.6848 & 0.6396 & 0.9059 & 0.8959 & 0.7986 & 0.4737 & 0.4805 \\
    \rowcolor[rgb]{ .949,  .949,  .949} PointMamba & Arxiv 2024 & \textcolor[rgb]{ 0,  .439,  .753}{\textbf{0.7787}} & \textcolor[rgb]{ 0,  .439,  .753}{\textbf{0.8204}} & 0.9163 & 0.9161 & 0.8983 & \textcolor[rgb]{ 0,  .439,  .753}{\textbf{0.6412}} & \textcolor[rgb]{ 1,  0,  0}{\textbf{0.7424}} \\
    Ours  & -     & \textcolor[rgb]{ 1,  0,  0}{\textbf{0.8074}} & \textcolor[rgb]{ 1,  0,  0}{\textbf{0.8555}} & \textcolor[rgb]{ 1,  0,  0}{\textbf{0.9468}} & \textcolor[rgb]{ 1,  0,  0}{\textbf{0.9469}} & \textcolor[rgb]{ 1,  0,  0}{\textbf{0.9692}} & \textcolor[rgb]{ 1,  0,  0}{\textbf{0.6678}} & \textcolor[rgb]{ 0,  .439,  .753}{\textbf{0.7417}} \\
    \bottomrule
    \bottomrule
    \end{tabular}}%
    \caption{Results on Anomaly-ShapeNet.}
  \label{tab:r1}%
\end{table}%

\end{appendices}

\end{document}